\def\year{2021}\relax
\documentclass[letterpaper]{article} 
\usepackage{aaai21}  
\usepackage{times}  
\usepackage{helvet} 
\usepackage{courier}  
\usepackage[hyphens]{url}  
\usepackage{graphicx} 
\usepackage{natbib}  
\usepackage{caption} 
\usepackage{xcolor}
\usepackage{enumitem}
\usepackage{amsmath,amssymb}
\usepackage{multirow}
\usepackage{boldline}
\usepackage{makecell}
\usepackage[switch]{lineno}
\title{Image-to-Image Retrieval \\ by Learning Similarity between Scene Graphs}
\author {

        Sangwoong Yoon\textsuperscript{\rm 1}\footnotemark,
        Woo Young Kang\textsuperscript{\rm 2},
        Sungwook Jeon\textsuperscript{\rm 3}, 
        SeongEun Lee\textsuperscript{\rm 3},
        Changjin Han\textsuperscript{\rm 3},\\
        Jonghun Park\textsuperscript{\rm 3}, and
        Eun-Sol Kim\textsuperscript{\rm 2}
        \\
        
}

\affiliations {
    \textsuperscript{\rm 1}Seoul National University Robotics Lab,
    \textsuperscript{\rm 2}Kakao Brain, 
    \textsuperscript{\rm 3}Seoul National University Information Management Lab\\
    \texttt{\small swyoon@robotics.snu.ac.kr, edwin.kang@kakaobrain.com, \{wookee3,ryuha96,changjin9653,jonghun\}@snu.ac.kr, eunsol.kim@kakaobrain.com}
}

\begin{document}

\maketitle
\renewcommand{\thefootnote}{\fnsymbol{footnote}}
\footnotetext[1]{Work done during an internship at Kakao Brain}
\renewcommand{\thefootnote}{\arabic{footnote}}
\begin{abstract}
As a scene graph compactly summarizes the high-level content of an image in a structured and symbolic manner, the similarity between scene graphs of two images reflects the relevance of their contents. 
Based on this idea, we propose a novel approach for image-to-image retrieval using scene graph similarity measured by graph neural networks.
In our approach, graph neural networks are trained to predict the proxy image relevance measure, computed from human-annotated captions using a pre-trained sentence similarity model.
We collect and publish the dataset for image relevance measured by human annotators to evaluate retrieval algorithms.
The collected dataset shows that our method agrees well with the human perception of image similarity than other competitive baselines.
\end{abstract}

\section{Introduction}

Image-to-image retrieval, the task of finding similar images to a query image from a database, is one of the fundamental problems in computer vision and is the core technology in visual search engines.
The application of image retrieval systems has been most successful in problems where each image has a clear representative object, such as landmark detection and instance-based retrieval \cite{gordo2016deep, mohedano2016bags, radenovic2016cnn}, or has explicit tag labels \cite{gong2014multi}.

\begin{figure}[t]
    \centering
    \includegraphics[width=0.40\textwidth,height=0.25\textheight]{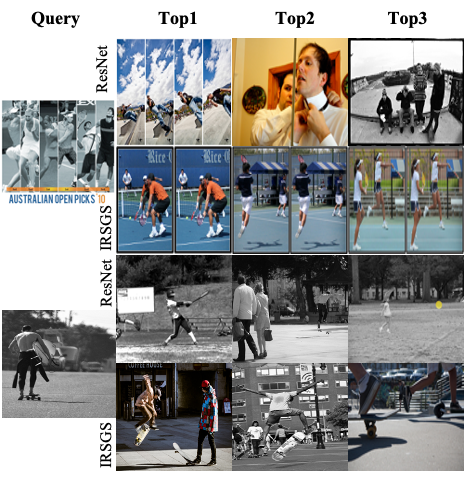}
    \caption{Image retrieval examples from ResNet and IRSGS. ResNet retrieves images with superficial similarity, e.g., grayscale or vertical lines, while IRSGS successfully returns images with correct context, such as playing tennis or skateboarding.}
    \label{fig:resnet-failure}
\end{figure}

However, performing image retrieval with complex images that have multiple objects and various relationships between them remains challenging for two reasons.
First, deep convolutional neural networks (CNNs), on which most image retrieval methods rely heavily, tend to be overly sensitive to low-level and local visual features \cite{zheng2017sift, zeiler2014visualizing, chen2018iterative}. 
As shown in Figure \ref{fig:resnet-failure}, nearest-neighbor search on ResNet-152 penultimate layer feature space returns images that are superficially similar but have completely different content.
Second, there is no publicly available labeled data to train and evaluate the image retrieval system for complex images, partly because quantifying similarity between images with multiple objects as label information is difficult.
Furthermore, a similarity measure for such complex images is desired to reflect semantics of images, i.e., the context and relationship of entities in images.

In this paper, we address these challenges and build an image retrieval system capable of finding semantically similar images to a query from a complex scene image database.
First of all, we propose a novel image retrieval framework, \textbf{Image Retrieval with Scene Graph Similarity} (\textbf{IRSGS}), which retrieves images with a similar scene graph to the scene graph of a query.
A scene graph represents an image as a set of objects, attributes, and relationships, summarizing the content of a complex image.
Therefore, the scene graph similarity can be an effective tool to measure semantic similarity between images.
IRSGS utilizes a graph neural networks to compute the similarity between two scene graphs, becoming more robust to confounding low-level features (Figure \ref{fig:resnet-failure}).

Also, we conduct a human experiment to collect human decisions on image similarity.
In the experiment, annotators are given a query image along with two candidate images and asked to select which candidate image is more similar to the query than the other.
With 29 annotators, we collect more than 10,000 annotations over more than 1,700 image triplets.
Thanks to the collected dataset, we can quantitatively evaluate the performance of image retrieval methods.
Our dataset is available online\footnote{https://github.com/swyoon/aaai2021-scene-graph-img-retr }.

However, it is costly to collect enough ground truth annotation from humans to supervise the image retrieval algorithm for a large image dataset, because the number of pairwise relationships to be labeled grows in $O(N^2)$ for the number of data $N$. 
Instead, we utilize human-annotated captions of images to define proxy image similarity, inspired by \citet{gordo2017beyond} which used term frequencies of captions to measure image similarity.
As a caption tends to cover important objects, attributes, and relationships between objects in an image, the similarity between captions is likely to reflect the contextual similarity between two images.
Also, obtaining captions is more feasible, as the number of the required captions grow in $O(N)$.
We use the state-of-the-art sentence embedding \cite{reimers2019sentence} method to compute the similarity between captions. The computed similarity is used to train a graph neural network in IRSGS and evaluate the retrieval results.

Tested on real-world complex scene images, IRSGS show higher agreement with human judgment than other competitive baselines. 
The main contributions of this paper can be summarized as follows:
\begin{itemize}[noitemsep]
    \item We propose IRSGS, a novel image retrieval framework that utilizes the similarity between scene graphs computed from a graph neural network to retrieve semantically similar images;
    \item We collect more than 10,000 human annotations for semantic-based image retrieval methods and publish the dataset into the public;
    \item We propose to train the proposed retrieval framework with the surrogate relevance measure obtained from image captions and a pre-trained language model;
    \item We empirically evaluate the proposed method and demonstrate its effectiveness over other baselines.
\end{itemize}

\begin{figure*}[ht]
\centering
\includegraphics[width=0.9\textwidth]{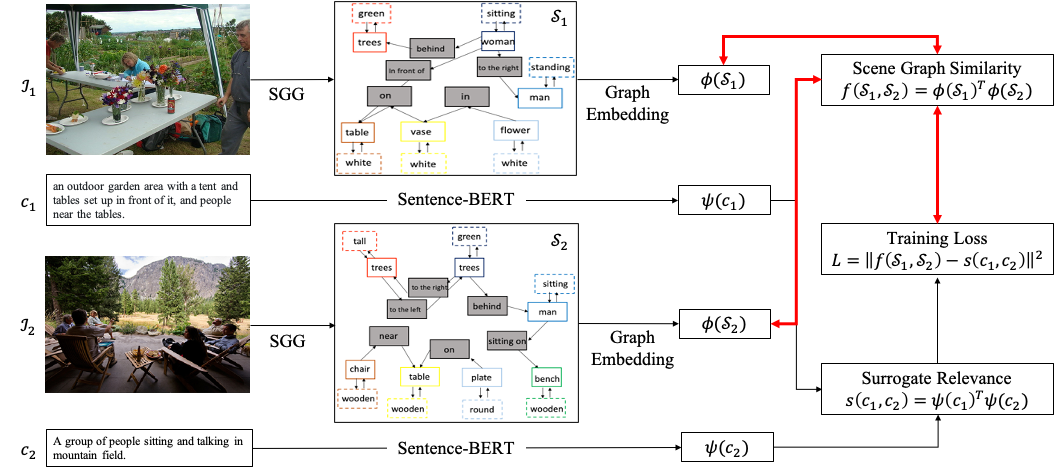}
\caption{An overview of IRSGS. Images $\mathcal{I}_1, \mathcal{I}_2$ are converted into vector representations $\phi(\mathcal{S}_1), \phi(\mathcal{S}_2)$ through scene graph generation (SGG) and graph embedding. The graph embedding function is learned to minimize mean squared error to surrogate relevance, i.e., the similarity between captions. The bold red bidirectional arrows indicate trainable parts. For retrieval, the learned scene graph similarity function is used to rank relevant images.
}
\label{fig:intro}
\end{figure*}

\section{Related Work}

\subsection{Image Retrieval}
Conventional image retrieval methods use visual feature representations, object categories, or text descriptions \cite{zheng2017sift,babenko2014neural,chen2019analysis,wei2016cross,Zhen_2019_CVPR,gu2017look,vo2019composing,gordo2017end}.
The activation of intermediate layers of CNN is shown to be effective as a representation of an image for image retrieval tasks. However, as shown in Figure \ref{fig:resnet-failure}, CNN often fails to capture semantic contents of images and is confounded by low-level visual features.

Image retrieval methods which reflects more semantic contents of images are investigated in \citet{gordo2017beyond,johnson2015image}.
\citet{gordo2017beyond} used term frequencies in regional captions to supervise CNN for image retrieval, but they did not utilize scene graphs. \citet{johnson2015image} proposed an algorithm retrieving images given a scene graph query. However, their approach does not employ graph-to-graph comparison and is not scalable.

\subsection{Scene Graphs}
A \textit{scene graph} \cite{johnson2015image} represents the content of an image in the form of a graph nodes of which represent objects, their attributes, and the relationships between them. 
After a large-scale real-world scene graph dataset manually annotated by humans in Visual Genome dataset \cite{krishna2017visual} was published, a number of applications such as image captioning \cite{wu2017image,lu2018neural,milewski2020scene} visual question answering \cite{teney2017graph}, and image-grounded dialog \cite{das2017visual} have shown the effectiveness of the scene graphs. 
Furthermore, various works, such as GQA\cite{hudson2018gqa}, VRD\cite{lu2016visual}, and VrR-VG\cite{liang2019vrr} provided the human-annotated scene graph datasets. 
Also, recent researches \cite{yang2018graph,xu2017scene,li2017scene} have suggested methods to generate scene graphs automatically.
Detailed discussion on scene graph generation will be made in Experimental Setup Section.

\subsection{Graph Similarity Learning} \label{sec:graph-similarity}

Many algorithms have been proposed for solving the isomorphism test or (sub-)graph matching task between two graphs. However, such methods are often not scalable to huge graphs or not applicable in the setting where node features are provided. Here, we review several state-of-the-art algorithms that are related to our application, image retrieval by graph matching. 
For the graph pooling perspective, we focus on two recent algorithms, the Graph Convolutional Network (GCN;\citet{kipf2016semi}) and the Graph Isomorphism Network (GIN;\cite{xu2018powerful}). 
GCN utilized neural network-based spectral convolutions in the Fourier domain to perform the convolution operation on a graph. 
GIN used injective aggregation and graph-level readout functions.
The learned graph representations, then, can be used to get the similarity of two graphs. 
Both networks transforms a graph into a fixed-length vector, enabling distance computation between two graphs in the vector space. 
Other studies viewed the graph similarity learning problem as the optimal transport problem \cite{solomon2016entropic,maretic2019got,alvarez2018gromov,xu2019scalable,xu2019gromov,titouan2019optimal}. 
Especially in Gromov Wasserstein Learning (GWL;\cite{xu2019gromov}), node embeddings were learned from associated node labels.
Thus the method can reflect not only a graph structure but also node features at the same time. 
Graph Matching Network (GMN;\cite{li2019graph}) used the cross-graph attention mechanism, which yields different node representations for different pairs of graphs.

\section{Image Retrieval with Scene Graph Similarity}

In this section, we describe our framework, Image Retrieval with Scene Graph Similarity (IRSGS). Given a query image, IRSGS first generates a query scene graph from the image and then retrieves images with a scene graph highly similar to the query scene graph. Figure \ref{fig:intro} illustrates the retrieval process. The similarity between scene graphs is computed through a graph neural network trained with surrogate relevance measure as a supervision signal.

\subsection{Scene Graphs and Their Generation}

Formally, a scene graph $\mathcal{S}=\{ \mathcal{O}, \mathcal{A}, \mathcal{R} \}$ of an image $\mathcal{I}$ is defined as a set of objects $\mathcal{O}$, attributes of objects $\mathcal{A}$ , and relations on pairs of objects $\mathcal{R}$. All objects, attributes, and relations are associated with a word label, for example, "car", "red", and "in front of". We represent a scene graph as a set of nodes and edges, i.e., a form of a conventional graph. All objects, attributes, and relations are treated as nodes, and associations among them are represented as undirected edges. Word labels are converted into 300-dimensional GloVe vectors \cite{pennington2014glove} and treated as node features.

Generating a scene graph from an image is equivalent to detecting objects, attributes, and relationships in the image. We employ a recently proposed method \cite{anderson2018bottom} in our IRSGS framework to generate scene graphs. While end-to-end training of scene graph generation module is possible in principle, a fixed pre-trained algorithm is used in our experiments to reduce the computational burden. We shall provide details of our generation process in Experimental Setup Section. Note that IRSGS is compatible with any scene graph generation algorithm and is not bound to the specific one we used in this paper.


\subsection{Retrieval via Scene Graph Similarity}

Given a query image $\mathcal{I}_q$, an image retrieval system ranks candidate images $\{\mathcal{I}_i\}_{i=1}^{N}$ according to the similarity to the query image $\text{sim}(\mathcal{I}_i, \mathcal{I}_q)$.
IRSGS casts this image retrieval task into a graph retrieval problem by defining the similarity between images as the similarity between corresponding scene graphs. Formally,
\begin{align}
    \text{sim}(\mathcal{I}_i, \mathcal{I}_j)=f(\mathcal{S}_i, \mathcal{S}_j)
\end{align}
where $\mathcal{S}_i, \mathcal{S}_j$ are scene graphs for $\mathcal{I}_i, \mathcal{I}_j$, respectively. We shall refer $f(\mathcal{S}_i, \mathcal{S}_j)$ as \textit{scene graph similarity}.

We compute the scene graph similarity from the inner product of two representation vectors of scene graphs. With a scene graph, a graph neural network is applied, and the resulting node representations are pooled to generate a unit $d$-dimensional vector $\phi=\phi(\mathcal{S}) \in \mathbb{R}^d$. The scene graph similarity is then given as follows:
\begin{align}
    f(\mathcal{S}_1, \mathcal{S}_2) = \phi(\mathcal{S}_1)^{\top}\phi(\mathcal{S}_2). \label{eq:inner}
\end{align}
We construct $\phi$ by computing the forward pass of graph neural networks to obtain node representations and then apply average pooling. 
We implement $\phi$ with either GCN or GIN, yielding two versions, IRSGS-GCN and IRSGS-GIN, respectively.

\subsection{Learning to Predict Surrogate Relevance}\label{sec:surrogate}

We define \textit{surrogate relevance measure} between two images as the similarity between their captions.
Let $c_i$ and $c_j$ are captions of image $\mathcal{I}_i$ and $\mathcal{I}_j$.
To compute the similarity between the captions, we first apply Sentence-BERT (SBERT; \citet{reimers2019sentence})\footnote{We use the code and the pre-trained model (bert-large-nli-mean-tokens) provided in\\ https://github.com/UKPLab/sentence-transformers.} and project the output to the surface of an unit sphere to obtain representation vectors $\psi(c_i)$ and $\psi(c_j)$. The surrogate relevance measure $s(c_i, c_j)$ is then given by their inner product: $s(c_i, c_j)=\psi(c_i)^\top \psi(c_j)$.
When there is more than one caption for an image, we compute the surrogate relevance of all caption pairs and take the average.
With the surrogate relevance, we are able to compute a proxy score for any pair of images in the training set, given their human-annotated captions.
To validate the proposed surrogate relevance measure, we collect human judgments of semantic similarity between images by conducting a human experiment (details in Human Annotation Collection Section).

We train the scene graph similarity $f$ by directly minimizing mean squared error from the surrogate relevance measure, formulating the learning as a regression problem. The loss function for $i$-th and $j$-th images is given as $L_{ij} =||f(\mathcal{S}_i, \mathcal{S}_j) - s(c_i, c_j) ||^2$.
Other losses, such as triplet loss or contrastive loss, can be employed as well. However, we could not find clear performance gains with those losses and therefore adhere to the simplest solution.

\begin{center} 
   \begin{table*}[ht]
   \centering
   \begin{small}
    \begin{tabular}{c|c|cccccc|c}
    \hlineB{2}
 \multicolumn{1}{c|}{\multirow{2}{*}{Method}} & \multicolumn{1}{c|}{\multirow{2}{*}{Data}} & \multicolumn{6}{c}{nDCG}  & \multicolumn{1}{|c}{\multirow{2}{*}{\begin{tabular}{@{}c@{}}Human\\ Agreement\end{tabular}}}\\ \cline{3-8}
 \multicolumn{1}{c|}{} & \multicolumn{1}{c|}{} & \multicolumn{1}{c}{5}  & \multicolumn{1}{c}{10} & \multicolumn{1}{c}{20} & \multicolumn{1}{c}{30} & \multicolumn{1}{c}{40} & \multicolumn{1}{c}{50} & \multicolumn{1}{|c}{}\\ \hline
 \multicolumn{1}{c|}{Inter Human} & \multicolumn{1}{c|}{-} & \multicolumn{1}{c}{-}  & \multicolumn{1}{c}{-} & \multicolumn{1}{c}{-} & \multicolumn{1}{c}{-} & \multicolumn{1}{c}{-} & \multicolumn{1}{c}{-} & \multicolumn{1}{|c}{0.730 $\pm$ 0.05}\\ 
 \multicolumn{1}{c|}{Caption SBERT} & \multicolumn{1}{c|}{Cap(HA)} & \multicolumn{1}{c}{1}  & \multicolumn{1}{c}{1} & \multicolumn{1}{c}{1} & \multicolumn{1}{c}{1} & \multicolumn{1}{c}{1} & \multicolumn{1}{c}{1} & \multicolumn{1}{|c}{0.700}\\
 \multicolumn{1}{c|}{Random} & \multicolumn{1}{c|}{-} & \multicolumn{1}{c}{0.136}  & \multicolumn{1}{c}{0.138} & \multicolumn{1}{c}{0.143} & \multicolumn{1}{c}{0.147} & \multicolumn{1}{c}{0.149} & \multicolumn{1}{c}{0.152} & \multicolumn{1}{|c}{0.472 $\pm$ 0.01}\\\hline
 \multicolumn{1}{c|}{Gen. Cap. SBERT} & \multicolumn{1}{c|}{Cap(Gen)} & \multicolumn{1}{c}{0.609}  & \multicolumn{1}{c}{0.628} & \multicolumn{1}{c}{0.657} & \multicolumn{1}{c}{0.681} & \multicolumn{1}{c}{0.703} & \multicolumn{1}{c}{0.726} & \multicolumn{1}{|c}{0.473}\\
 \multicolumn{1}{c|}{ResNet} & \multicolumn{1}{c|}{I} &  \multicolumn{1}{c}{0.687}  & \multicolumn{1}{c}{0.689} & \multicolumn{1}{c}{0.691} & \multicolumn{1}{c}{0.692} & \multicolumn{1}{c}{0.693} & \multicolumn{1}{c}{0.693} & \multicolumn{1}{|c}{0.494}\\
 \multicolumn{1}{c|}{ResNet-FT} & \multicolumn{1}{c|}{I} & \multicolumn{1}{c}{0.642}  & \multicolumn{1}{c}{0.656} & \multicolumn{1}{c}{0.682} & \multicolumn{1}{c}{0.703} & \multicolumn{1}{c}{0.724} & \multicolumn{1}{c}{0.745} & \multicolumn{1}{|c}{0.478}\\ \hline
 \multicolumn{1}{c|}{Object Count}  & \multicolumn{1}{c|}{I+SG} & \multicolumn{1}{c}{0.736}  & \multicolumn{1}{c}{0.749} & \multicolumn{1}{c}{0.770} & \multicolumn{1}{c}{0.788} & \multicolumn{1}{c}{0.804} & \multicolumn{1}{c}{0.819} & \multicolumn{1}{|c}{0.587}\\
 \multicolumn{1}{c|}{GMN}   & \multicolumn{1}{c|}{I+SG} & \multicolumn{1}{c}{0.721}  & \multicolumn{1}{c}{0.735} & \multicolumn{1}{c}{0.755} & \multicolumn{1}{c}{0.771} & \multicolumn{1}{c}{0.786} & \multicolumn{1}{c}{0.801} & \multicolumn{1}{|c}{0.535}\\
 \multicolumn{1}{c|}{IRSGS-GIN}   & \multicolumn{1}{c|}{I+SG} & \multicolumn{1}{c}{0.751}  & \multicolumn{1}{c}{0.768} & \multicolumn{1}{c}{0.790} & \multicolumn{1}{c}{0.808} & \multicolumn{1}{c}{0.824} & \multicolumn{1}{c}{0.839} & \multicolumn{1}{|c}{0.576}\\
 \multicolumn{1}{c|}{IRSGS-GCN}  & \multicolumn{1}{c|}{I+SG} & \multicolumn{1}{c}{\textbf{0.784}}  & \multicolumn{1}{c}{\textbf{0.795}} & \multicolumn{1}{c}{\textbf{0.814}} & \multicolumn{1}{c}{\textbf{0.829}} & \multicolumn{1}{c}{\textbf{0.844}} & \multicolumn{1}{c}{\textbf{0.856}} & \multicolumn{1}{|c}{\textbf{0.602}}\\
 \hlineB{2}
\end{tabular}    
\end{small}
\caption{Image retrieval results on VG-COCO with human-annotated scene graphs. Data column indicates which data modalities are used. Cap(HA): human-annotated captions. Cap(Gen): machine-generated captions. I: image. SG: scene graphs.}
\label{result:vg-coco-gt}
\end{table*}
\end{center}

\section{Human Annotation Collection}\label{sec:human-label}

We collect semantic similarity annotations from humans to validate the proposed surrogate relevance measure and to evaluate image retrieval methods.
Through our web-based annotation system, a human labeler is asked whether two candidate images are semantically similar to a given query image. The labeler may choose one of four answers: either of the two candidate images is more similar than the other, images in the triplet are semantically identical, or neither of the candidate images is relevant to the query.
We collect 10,712 human annotations from 29 human labelers for 1,752 image triplets constructed from the test set of the VG-COCO, the dataset we shall define in Experimental Setup Section.

A query image of a triplet is randomly selected from the query set defined in the following section. Two candidate images are randomly selected from the rest of the test set, subjected to two constraints. First, the rank of a candidate image should be less than or equal to 100 when the whole test set is sorted according to cosine similarity in ResNet-152 representation to the query image. Second, the surrogate relevance of a query-candidate image pair in a triplet should be larger than the other, and the difference should be greater than 0.1. This selection criterion produces visually close yet semantically different image triplets.

We define the \emph{human agreement score} to measure the agreement between decisions of an algorithm and that of the human annotators, in a similar manner presented in \cite{gordo2017beyond}. The score is an average portion of human annotators who made the same decision per each triplet. Formally, given a triplet, let $s_1$ (or $s_2$) be the number of human annotators who chose the first (or the second) candidate image is more semantically similar to the query, $s_3$ be the number of annotators who answered that all three images are identical, and $s_4$ be the number of annotators who marked the candidates as irrelevant. If an algorithm choose either one of candidate images is more relevant, the human agreement score for a triplet is $\frac{s_i + 0.5s_3}{s_1 + s_2 + s_3 + s_4}$, where $i=1$ if the algorithm determines that the first image is semantically closer and $i=2$ otherwise. The score is averaged over triplets with $s_1 + s_2 \geq 2$. Randomly selecting one of two candidate images produces an average human agreement of 0.472 with a standard deviation of 0.01. Note that the agreement of random decision is lower than 0.5 due to the existence of the human choice of "both" ($s_3$) and "neither" ($s_4$).

The alignment between labelers is also measured with the human agreement score in a leave-one-out fashion. If a human answers that both candidate images are relevant, the score for the triplet is $\frac{0.5s_1 + 0.5s_2 + s_3}{s_1 + s_2 + s_3 + s_4}$, where $s_1\dots s_4$ are computed from the rest of annotators. If a human marks that neither of the candidates is relevant for a triplet, the triplet is not counted in the human agreement score. The mean human agreement score among those annotators is 0.727, and the standard deviation is 0.05. We will make the human annotation dataset public after the review.

\section{Experimental Setup}

\subsection{Data} \label{sec:data}
In experiments, we use two image datasets involving diverse semantics. The first dataset is the intersection of the Visual Genome \cite{krishna2017visual} and MS-COCO \cite{lin2014microsoft}, which we will refer to as VG-COCO. In VG-COCO, each image has a scene graph annotation provided by Visual Genome and five captions provided by MS-COCO. We utilize the refined version of scene graphs provided by \cite{xu2017scene} and their train-test split. After removing the images with empty scene graphs, we obtain fully annotated 35,017 training images and 13,203 test images. We randomly select a fixed set of 1,000 images among the test set and define them as a \textit{query set}. For each query image, a retrieval algorithm is asked to rank the other 13,202 images in the test set according to the semantic similarity. Besides the annotated scene graphs, we automatically generate scene graphs for all images and experiment with our approach to both human-labeled and machine-generated scene graphs.

The second dataset is Flickr30K \cite{flickrentitiesijcv}, where five captions are provided per an image. Flickr30K contains 30,000 training images, 1,000 validation images, and 1,000 testing images. For Flickr30k, the whole test set is the query set. During the evaluation, an algorithm ranks the other 999 images given a query image in a test set. Scene graphs are generated in the same manner as in the VG-COCO dataset.


\subsection{Scene Graph Generation Detail}\label{sec:sgg}
Since we focus on learning graph embeddings when two scene graphs are given for the image-to-image retrieval task, we use the conventional scene graph generation process. 
Following the works \cite{anderson2018bottom}, objects in images are detected by Faster R-CNN method, and the name and attributes of the objects are predicted based on the ResNet-101 features from the detected bounding boxes.
We keep up to 100 objects with a confidence threshold of 0.3.
To predict relation labels between objects after extracting information about the objects, we used the frequency prior knowledge constructed from the GQA dataset that covers 309 kinds of relations.\footnote{We have been tried to predict relation labels by using recently suggested SGG algorithms, such as \cite{yang2018graph,xu2017scene,li2017scene}. However, we could not achieve any improvement in image retrieval tasks. The reasons might be that 1)  small size vocabularies for object and relation are used for the conventional SGG setting (only 150/50 kinds of objects/relations), 2) the algorithms do not predict the attributes, and 3) the annotated scene graphs used for training the methods have very sparse relations. } 
For each pair of the detected objects, relationships are predicted based on the frequency prior with confidence threshold 0.2.
To give position-specific information, the coordinates of the detected bbox are used.
Here, we should note that even though the suggested method to generate a scene graph is quite simple than other methods \cite{yang2018graph,xu2017scene,li2017scene}, it outperforms all the others.

\subsection{Two-Step Retrieval using Visual Features}

In information retrieval, it is a common practice to take a two-step approach \cite{wang2019enhancing,bai2016sparse}: retrieving roughly relevant items first and then sorting (or "re-ranking") the retrieved items according to the relevance. We also employ this approach in our experiment. For a query image, we first retrieve $K$ images that are closest to the query in a ResNet-152 feature representation space formed by the 2048-dimension activation vector of the last hidden layer. The distance is measured in cosine similarity. This procedure generates a set of good candidate images which have a high probability of having strong semantic similarity. This approximate retrieval step can be further boosted by using an approximate nearest neighbor engine such as Faiss \cite{JDH17} and is critical if the following re-ranking step is computationally involved. We use this approximate pre-ranking for all experiments with $K=100$ unless otherwise mentioned. Although there is large flexibility of designing this step, we shall leave other possibilities for future exploration as the re-ranking step is our focus.

\subsection{Training Details}

We use Adam optimizer with the initial learning rate of 0.0001. We multiply 0.9 to the learning rate every epoch. We set batch size as 32, and models are trained for 25 epochs.
In each training step, a mini-batch of pairs is formed by randomly drawing samples. When drawing the second sample in a pair, we employ an oversampling scheme to reinforce the learning of pairs with large similarity values.
With a probability of 0.5, the second sample in a pair is drawn from 100 most relevant samples with the largest surrogate relevance score to the first sample. Otherwise, we select the second sample from the whole training set. 
Oversampling improves both quantitative and qualitative results and is apply identically for all methods except for GWL where the scheme is not applicable.

\begin{center}  
   \begin{table}[t]
   \setlength\tabcolsep{2.5pt}
   \centering
   \begin{small}
    \begin{tabular}{c|cccc|c}
    \hlineB{2}
 \multicolumn{1}{c|}{\multirow{2}{*}{Method}} & \multicolumn{4}{c}{nDCG}  & \multicolumn{1}{|c}{\multirow{2}{*}{\begin{tabular}{@{}c@{}}Human\\ Agreement\end{tabular} }}\\ \cline{2-5}
 \multicolumn{1}{c|}{} & \multicolumn{1}{c}{5}  & \multicolumn{1}{c}{10} & \multicolumn{1}{c}{20} & \multicolumn{1}{c}{40} & \multicolumn{1}{|c}{}\\ \hline
 \multicolumn{1}{c|}{Inter Human} & \multicolumn{1}{c}{-}  & \multicolumn{1}{c}{-} & \multicolumn{1}{c}{-} & \multicolumn{1}{c}{-} & \multicolumn{1}{|c}{0.730}\\ 
 \multicolumn{1}{c|}{Caption SBERT} & \multicolumn{1}{c}{1}  & \multicolumn{1}{c}{1} & \multicolumn{1}{c}{1} & \multicolumn{1}{c}{1} & \multicolumn{1}{|c}{0.700}\\
 \multicolumn{1}{c|}{Random} & \multicolumn{1}{c}{0.136}  & \multicolumn{1}{c}{0.138} & \multicolumn{1}{c}{0.143} & \multicolumn{1}{c}{0.149}  & \multicolumn{1}{|c}{0.472}\\\hline
 \multicolumn{1}{c|}{Gen. Cap. SBERT} & \multicolumn{1}{c}{0.609}  & \multicolumn{1}{c}{0.628} & \multicolumn{1}{c}{0.657} & \multicolumn{1}{c}{0.703}  & \multicolumn{1}{|c}{0.473}\\
 \multicolumn{1}{c|}{ResNet} & \multicolumn{1}{c}{0.687}  & \multicolumn{1}{c}{0.689} & \multicolumn{1}{c}{0.691} & \multicolumn{1}{c}{0.693}  & \multicolumn{1}{|c}{0.494}\\
 \multicolumn{1}{c|}{ResNet-FT} & \multicolumn{1}{c}{0.642}  & \multicolumn{1}{c}{0.656} & \multicolumn{1}{c}{0.682} & \multicolumn{1}{c}{0.724}  & \multicolumn{1}{|c}{0.478}\\ \hline
 \multicolumn{1}{c|}{Object Count} & \multicolumn{1}{c}{0.73}  & \multicolumn{1}{c}{0.743} & \multicolumn{1}{c}{0.761} & \multicolumn{1}{c}{0.794}  & \multicolumn{1}{|c}{0.581}\\
 \multicolumn{1}{c|}{GWL} & \multicolumn{1}{c}{0.748}  & \multicolumn{1}{c}{0.758} & \multicolumn{1}{c}{0.774} & \multicolumn{1}{c}{0.803}  & \multicolumn{1}{|c}{0.598}\\
 \multicolumn{1}{c|}{GMN} & \multicolumn{1}{c}{0.728}  & \multicolumn{1}{c}{0.740} & \multicolumn{1}{c}{0.755} & \multicolumn{1}{c}{0.781}  & \multicolumn{1}{|c}{0.539}\\
 \multicolumn{1}{c|}{IRSGS-GIN} & \multicolumn{1}{c}{0.764}  & \multicolumn{1}{c}{0.781} & \multicolumn{1}{c}{0.802} & \multicolumn{1}{c}{0.834}  & \multicolumn{1}{|c}{\textbf{0.612}}\\ 
 \multicolumn{1}{c|}{IRSGS-GCN} & \multicolumn{1}{c}{\textbf{0.771}}  & \multicolumn{1}{c}{\textbf{0.784}} & \multicolumn{1}{c}{\textbf{0.805}} & \multicolumn{1}{c}{\textbf{0.836}}  & \multicolumn{1}{|c}{0.611}\\
 \hlineB{2}
\end{tabular}
\end{small}
\caption{Image retrieval results on VG-COCO with machine-generated scene graphs. Baselines which do not use scene graphs are identical to the corresponding rows of Table \ref{result:vg-coco-gt}.}
   \label{result:vg-coco-gen}
\end{table}
\end{center}

\section{Experiments}

\subsection{Evaluation} \label{sec:eval}

We benchmark IRSGS and other baselines with VG-COCO and Flickr30K.
Images in the query set are presented as queries, and the relevance of the images ranked by an image retrieval algorithm is evaluated with two metrics. 
First, we compute normalized discounted cumulative gain (nDCG) with the surrogate relevance as gain. 
A larger nDCG value indicates stronger enrichment of relevant images in the retrieval result. 
In nDCG computation, surrogate relevance is clipped at zero to ensure its positivity.
Second, the agreement between a retrieval algorithm and decision of human annotators is measured in a method described in Human Annotation Collection Section.

\subsection{Baseline Methods}
\textbf{ResNet-152 Features}
Image retrieval is performed based on the cosine similarity in the last hidden representation of ResNet-152 pre-trained on ImageNet.

\textbf{Generated Caption} 
To test whether machine-generated captions can be an effective means for semantic image retrieval, we generate captions of images by soft attention model \cite{xu2015show} pretrained on Flickr30k dataset \cite{flickrentitiesijcv}. 
We obtain SBERT representations of generated captions, and their cosine similarity is used to perform image retrieval.

\textbf{Object Count (OC)} Ignoring relation information given in a scene graph, we transform a scene graph into a vector of object counts. Then, we compute the cosine similarity of object count vectors to perform image retrieval.

\textbf{ResNet Finetune (ResNet-FT)} We test whether a ResNet-152 can be fine-tuned to capture semantic similarity.
Similarly to Siamese Network \cite{bromley1994signature}, ResNet feature extractor is trained to produce cosine similarity between images close to their surrogate relevance measure.

\textbf{Gromov-Wasserstein Learning (GWL)}
Based on Gromov-Wasserstein Learning (GWL) framework \cite{xu2019gromov}, we obtain a transport map using a proximal gradient method \cite{xie2018fast}.
A transport cost, a sum of Gromov-Wasserstein discrepancy and Wasserstein discrepancy, is calculated with the transport map and the cost matrix, and used for retrieval.
The method is computationally demanding, and we only tested the method for VG-COCO with generated scene graphs setting in Table \ref{result:vg-coco-gen}.

\textbf{Graph Matching Networks (GMN)} GMNs are implemented based on the publicly available code\footnote{https://github.com/deepmind/deepmind-research/tree/master/\\graph\_matching\_networks}. We use four propagation layers with shared weights. The propagation in the reverse direction is allowed, and the propagated representation is updated using the gated recurrent unit.
Final node representations are aggregated by summation, resulting in a 128-dimensional vector which is then fed to a multi-layer perceptron to produce final scalar output. As GMN is capable of handling edge features, we leave relations as edges instead of transforming them as nodes. To indicate object-attribute connections, we append additional dimensionality to edge feature vectors and define a feature vector of an edge between an object and an attribute is a one-hot vector where only the last dimension is non-zero.

\subsection{Graph Embedding Methods in IRSGS}

Here, we describe implementation details of graph neural networks used in IRSGS.

\textbf{IRSGS-GCN}
A scene graph is applied with GCN and the final node representations are aggregated via mean pooling and scaled to the unit norm, yielding a representation vector $\phi(\mathcal{S})$. 
We use three graph convolution layers with 300 hidden neurons in each layer. 
The first two layers are followed by ReLU nonlinearity. 
Stacking more layers does not introduce clear improvement. 
We always symmetrize the adjacency matrix before applying GCN.

\textbf{IRSGS-GIN}
Similarly to GCN, we stack three GIN convolution layers with 300 hidden neurons in each layer. For multi-layer perceptrons required for each layer, we use one hidden layer with 512 neurons with ReLU nonlinearity. Other details are the same as that of the GCN case.

\begin{figure*}[ht]
\centering
\includegraphics[width=0.9\textwidth]{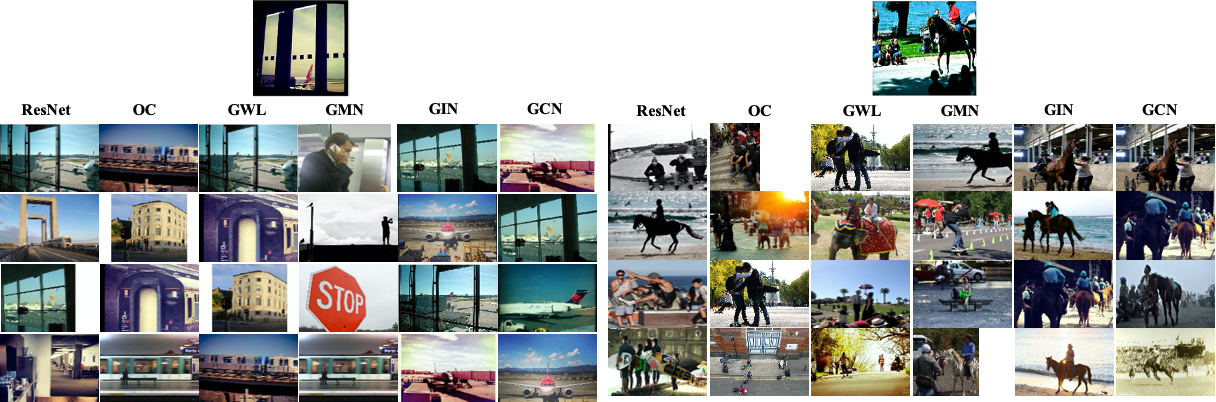}
\caption{Four most similar images retrieved by six algorithms. OC: Object Count, GIN: IRSGS-GIN, GCN: IRSGS-GCN. The visual genome ids for the query images are 2323522 and 2316427.}\label{fig:qual_result1}
\end{figure*}



\begin{center}
   \begin{table}[t]
   \centering
   \begin{small}
    \begin{tabular}{c|cccc}
    \hlineB{2}
 \multicolumn{1}{c|}{\multirow{2}{*}{Method}} & \multicolumn{4}{c}{nDCG}  \\ 
 \cline{2-5}
 \multicolumn{1}{c|}{} & \multicolumn{1}{c}{5}  & \multicolumn{1}{c}{10} & \multicolumn{1}{c}{20} & \multicolumn{1}{c}{40}  \\ \hline
 \multicolumn{1}{c|}{Captions SBERT} & \multicolumn{1}{c}{1}  & \multicolumn{1}{c}{1} & \multicolumn{1}{c}{1} & \multicolumn{1}{c}{1}  \\
 \multicolumn{1}{c|}{Random} & \multicolumn{1}{c}{0.195}  & \multicolumn{1}{c}{0.209} & \multicolumn{1}{c}{0.223} &  \multicolumn{1}{c}{0.245}  \\\hline
 \multicolumn{1}{c|}{Gen. Cap. SBERT} & \multicolumn{1}{c}{0.556}  & \multicolumn{1}{c}{0.576} & \multicolumn{1}{c}{0.610} & \multicolumn{1}{c}{0.659} \\
 \multicolumn{1}{c|}{Resnet} & \multicolumn{1}{c}{0.539}  & \multicolumn{1}{c}{0.541} & \multicolumn{1}{c}{0.541}  & \multicolumn{1}{c}{0.542}  \\ 
  \multicolumn{1}{c|}{ResNet-FT} & \multicolumn{1}{c}{0.368}  & \multicolumn{1}{c}{0.393} & \multicolumn{1}{c}{0.433} & \multicolumn{1}{c}{0.502}  \\\hline
 \multicolumn{1}{c|}{Object Count} & \multicolumn{1}{c}{0.511}  & \multicolumn{1}{c}{0.530} & \multicolumn{1}{c}{0.560} & \multicolumn{1}{c}{0.615}  \\
 \multicolumn{1}{c|}{IRSGS-GIN} & \multicolumn{1}{c}{0.564}  & \multicolumn{1}{c}{0.584} & \multicolumn{1}{c}{0.618} & \multicolumn{1}{c}{\textbf{0.673}}  \\
 \multicolumn{1}{c|}{IRSGS-GCN} & \multicolumn{1}{c}{\textbf{0.567}}  & \multicolumn{1}{c}{\textbf{0.590}} & \multicolumn{1}{c}{\textbf{0.623}} & \multicolumn{1}{c}{0.672}  \\
  \hlineB{2}
    \end{tabular}
\end{small}
\caption{Image retrieval results on Flickr30K with machine-generated scene graphs.}
   \label{result:f30k-gen}
\end{table}
\end{center}

\subsection{Quantitative Results} \label{subsec:quant_results}
From Table \ref{result:vg-coco-gt}, Table \ref{result:vg-coco-gen}, and Table \ref{result:f30k-gen}, IRSGS shows larger nDCG score than baselines across datasets (VG-COCO and Flickr30K) and methods of obtaining scene graphs (human-annotated and machine-generated).
IRSGS also achieves best agreement to human annotator's perception on semantic similarity, as it can be seen from Table \ref{result:vg-coco-gt} and Table \ref{result:vg-coco-gen}.

Comparing Table \ref{result:vg-coco-gt} and Table \ref{result:vg-coco-gen}, we found that using machine-generated scene graphs instead of human-annotated ones does not deteriorate the retrieval performance.
This result shows that IRSGS does not need human-annotated scene graphs to perform successful retrieval and can be applied to a dataset without scene graph annotation.
In fact, Flickr30K is the dataset without scene graph annotation, and IRSGS still achieves excellent retrieval performance in Flickr30K with machine-generated scene graphs.

On the other hand, using machine-generated captions in retrieval results in significantly poor nDCG scores and human agreement scores.
Unlike human-annotated captions, machine-generated captions are crude in quality and tend to miss important details of an image.
We suspect that scene graph generation is more stable than caption generation since it can be done in a systematic manner, i.e., predicting objects, attributes, and relations in a sequential way.

While not showing the optimal performance, GWL and GMN also show competitive performance over other methods based on generated captions and ResNet.
This overall tendency of competence of graph-based method is interesting and implies the effectiveness of scene graphs in capturing semantic similarity between images.

Note that in Caption SBERT, retrieval is performed with surrogate relevance, and their human agreement scores indicate the agreement between surrogate relevance and human annotations.
With the highest human agreement score than any other algorithms, this result assures that the proposed surrogate relevance reflects the human perception of semantic similarity well.

\subsection{Qualitative Results} \label{sec:quant_result}
Figure \ref{fig:resnet-failure} and Figure \ref{fig:qual_result1} show the example images retrieved from the retrieval methods we test.
Pitfalls of baseline methods that are not based on scene graphs can be noted.
As mentioned in Introduction, retrieval with ResNet features often neglects the semantics and focuses on the superficial visual characteristics of images.
On the contrary, OC only accounts for the presence of objects, yielding images with misleading context. For example, in the left panel of Figure \ref{fig:qual_result1}, OC simply returns images with many windows.
IRSGS could retrieve images containing similar objects with similar relations to the query image, for example, an airplane on the ground, or a person riding a horse.

\section{Discussion}

\textbf{Ablation Study} We also perform an ablation experiment for effectiveness of each scene graph component (Table \ref{result:vg-ablation}).
In this experiment, we ignore attributes or randomize relation information from IRSGS-GCN framework. In both cases, nDCG and Human agreement scores are higher than the Object Count that uses only object information. This indicates that both attributes and relation information are useful to improve the image retrieval performance of the graph matching-based algorithm. Further, randomizing relations drops performance more than ignoring attribute information, which means that relations are important for capturing the human perception of semantic similarity.

\textbf{Comparison to \citet{johnson2015image}} We exclude \citet{johnson2015image} from our experiment because the CRF-based algorithm from \citet{johnson2015image} is not feasible in a large-scale image retrieval problem. One of our goals is to tackle a large-scale retrieval problem where a query is compared against more than ten thousand images. Thus, we mainly consider methods that generate a compact vector representation of an image or a scene graph (Eq.(2)). However, the method in \citet{johnson2015image} requires object detection results to be additionally stored and extra computation for all query-candidate pairs to be done in the retrieval phase. Note that \citet{johnson2015image} only tested their algorithm on 1,000 test images, while we benchmark algorithms using 13,203 candidate images.  

\textbf{Effectiveness of Mean Pooling and Inner Product}
One possible explanation for the competitive performance of IRSGS-GCN and IRSGS-GIN is that the mean pooling and inner product are particularly effective in capturing similarity between two sets.
Given two sets of node representations $\{a_1,\cdots,a_N\}$ and $\{b_1,\cdots,b_M\}$, the inner product of their means are given as $\sum_{i,j} a_i^\top b_j / (NM)$, the sum of the inner product between all pairs. 
This expression is proportional to the number of common elements in the two sets, especially when $a_i^\top b_j$ is 1 if $a_i = b_j$ and 0 otherwise, measuring the similarity between the two sets.
If the inner product values are not binary, then the expression measures the set similarity in a "soft" way.

\begin{center}
   \begin{table}[t]
   \setlength\tabcolsep{2.5pt}
   \centering
   \begin{small}
    \begin{tabular}{c|cccc|c}
    \hlineB{2}
 \multicolumn{1}{c|}{\multirow{2}{*}{Method}} & \multicolumn{4}{c}{nDCG}  & \multicolumn{1}{|c}{\multirow{2}{*}{\begin{tabular}{@{}c@{}}Human\\ Agreement\end{tabular}  }}\\ \cline{2-5}
 \multicolumn{1}{c|}{} & \multicolumn{1}{c}{5}  & \multicolumn{1}{c}{10} & \multicolumn{1}{c}{20} & \multicolumn{1}{c}{40} & \multicolumn{1}{|c}{}\\ \hline
 \multicolumn{1}{c|}{IRSGS-GCN} & \multicolumn{1}{c}{0.771}  & \multicolumn{1}{c}{0.784} & \multicolumn{1}{c}{0.805} & \multicolumn{1}{c}{0.836} & \multicolumn{1}{|c}{0.611}\\
 \multicolumn{1}{c|}{No Attribute} & \multicolumn{1}{c}{0.767}  & \multicolumn{1}{c}{0.782} & \multicolumn{1}{c}{0.803} & \multicolumn{1}{c}{0.834}  & \multicolumn{1}{|c}{0.606}\\
 \multicolumn{1}{c|}{Random Relation} & \multicolumn{1}{c}{0.764}  & \multicolumn{1}{c}{0.777} &
 \multicolumn{1}{c}{0.797} & \multicolumn{1}{c}{0.828}  & \multicolumn{1}{|c}{0.604}\\
 \multicolumn{1}{c|}{Object Count} & \multicolumn{1}{c}{0.730}  & \multicolumn{1}{c}{0.743} & \multicolumn{1}{c}{0.761} & \multicolumn{1}{c}{0.794}  & \multicolumn{1}{|c}{0.581}\\ \hlineB{2}
    \end{tabular}
\end{small}
\caption{Scene graph component ablation experiment results on VG-COCO. Machine-generated scene graphs are used.}
   \label{result:vg-ablation}
\end{table}
\end{center}

\section{Conclusion}
In this paper, we tackle the image retrieval problem for complex scene images where multiple objects are present in various contexts.
We propose IRSGS, a novel image retrieval framework, which leverages scene graph generation and a graph neural network to capture semantic similarity between complex images.
IRSGS is trained to approximate surrogate relevance measure, which we define as a similarity between captions.
By collecting real human data, we show that both surrogate relevance and IRSGS show high agreement to human perception on semantic similarity.
Our results show that an effective image retrieval system can be built by using scene graphs with graph neural networks.
As both scene graph generation and graph neural networks are techniques that are rapidly advancing, we believe that the proposed approach is a promising research direction to pursue.

\section*{Acknowledgements}
Sangwoong Yoon is partly supported by the National Research Foundation of Korea Grant (NRF/MSIT2017R1E1A1A03070945) and MSIT-IITP (No. 2019-0-01367, BabyMind).

\bibliography{aaai21}

\section{Appendix}
\subsection{Computational Property}
lIRSGS is scalable in terms of both computing time and memory, adding only marginal overhead over a conventional image retrieval system.
For candidate images in a database, their graph embeddings and ResNet features are pre-computed and stored.
Generating a scene graph for a query image is mainly based on the object detection which can be run almost in real-time.
Searching over the database is essentially a nearest neighbor search, which is fast for the small ($<$ 100,000 images) number of images, and can be accelerated for a larger database with an approximate nearest neighbor search engines, such as Faiss \cite{JDH17}.
On the contrary, algorithms which use explicit graph matching, such as GWL and GMN, are significantly less scalable than IRSGS, because representation vectors from those methods cannot be pre-computed.
Given a generated scene graph, processing a pair of images takes approximately 15 seconds and 0.002 seconds for GWL and GMN, respectively. When retrieving from a database of 10,000 images, 0.002 seconds for a pair results in 20 seconds per a query, not applicable for a practical retrieval system.
On the other hand, IRSGS takes less than 0.001 seconds per a pair of images when the graph embeddings are not pre-computed and is more than 10 times faster when the embeddings are pre-computed and only the inner products to the query are computed.

\subsection{Two-Stage Retrieval}
The initial retrieval using ResNet is beneficial in two aspects: retrieval quality and speed. ResNet-based retrieval indeed introduces the bias but in a good way; the ResNet-based stage increases human agreement for all retrieval methods, possibly by excluding visually irrelevant images. Some baselines, such as graph matching networks, are not computationally feasible without the initial retrieval. However, IRSGS is computationally feasible without ResNet-based retrieval because the representations of images can be pre-computed and indexed. We empirically found that k=100 showed a good trade-off between computational cost and performance.

\subsection{Comparison to SPICE}
We initially excluded SPICE\cite{anderson2016spice} from experiments not because of its computational property but because of the exact matching mechanism that SPICE is based on. By definition, SPICE would consider two semantically similar yet distinct words as different. Meanwhile, IRSGS is able to match similar words since it utilizes the continuous embeddings of words. Still, SPICE can be an interesting baseline, and we will consider adding it for comparison.

\subsection{Full Resolution Figures}

Here, we provide figures presented in the main manuscript in their full scale.

\begin{figure}
    \centering
    \includegraphics[width=0.40\textwidth]{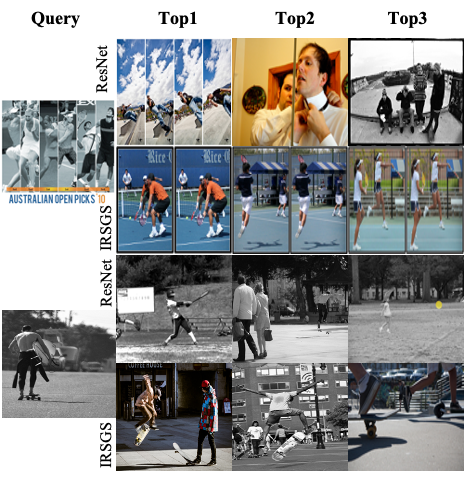}
    \caption{Image retrieval examples from ResNet and IRSGS. ResNet retrieves images with superficial similarity, e.g., grayscale or vertical lines, while IRSGS successfully returns images with correct context, such as playing tennis or skateboarding.}
    \label{fig:resnet-failure-full}
\end{figure}

\begin{figure*}[ht]
\centering
\includegraphics[width=\textwidth]{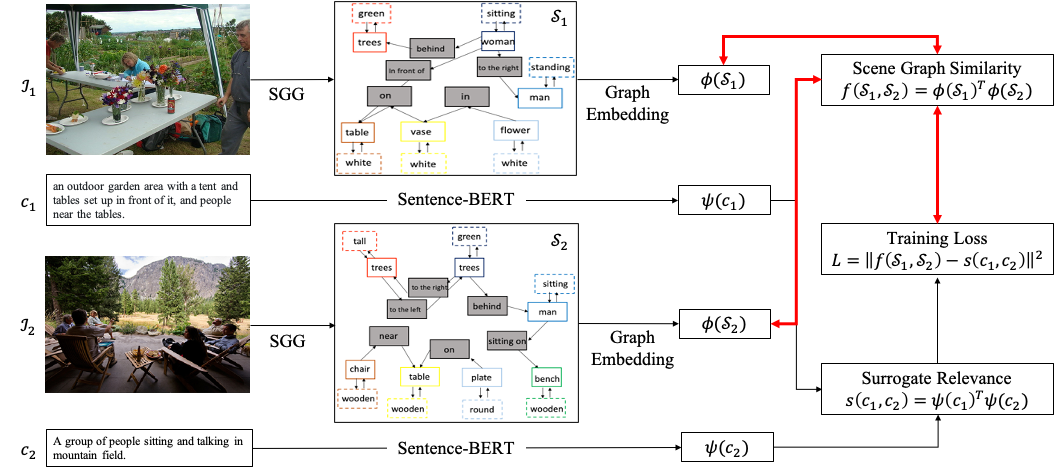}
\caption{An overview of IRSGS. Images $\mathcal{I}_1, \mathcal{I}_2$ are converted into vector representations $\phi(\mathcal{S}_1), \phi(\mathcal{S}_2)$ through scene graph generation (SGG) and graph embedding. The graph embedding function is learned to minimize mean squared error to surrogate relevance, i.e., the similarity between captions. The bold red bidirectional arrows indicate trainable parts. For retrieval, the learned scene graph similarity function is used to rank relevant images.
}
\label{fig:intro-full}
\end{figure*}

\begin{figure*}[!t]
\centering
\includegraphics[width=0.8\textwidth]{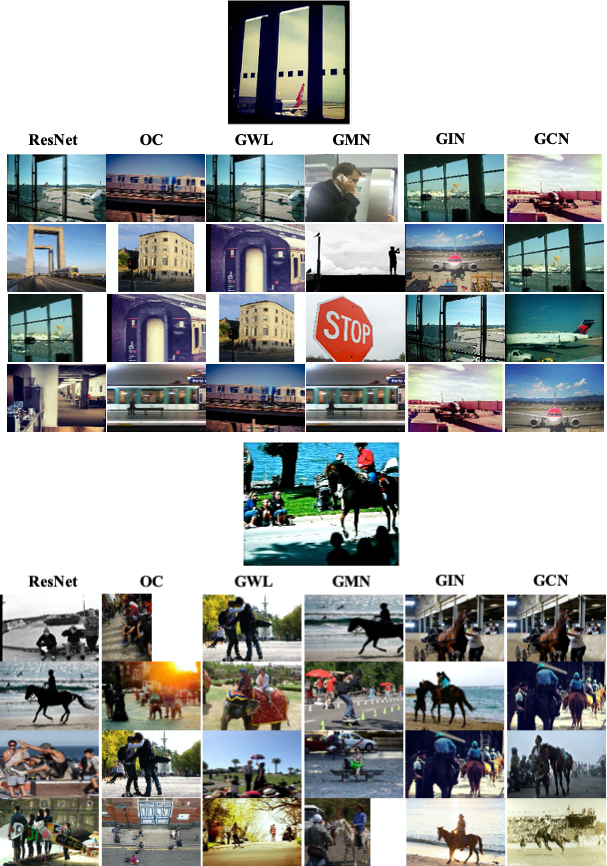}
\caption{Four most similar images retrieved by six algorithms. OC: Object Count, GIN: IRSGS-GIN, GCN: IRSGS-GCN. The visual genome ids for the query images are 2323522 and  2316427.}\label{fig:qual_result1-full}
\end{figure*}
\end{document}